\documentclass{article}

\usepackage[final,nonatbib]{neurips_2022}

\usepackage[numbers]{natbib}

\usepackage[utf8]{inputenc} 
\usepackage[T1]{fontenc}    
\usepackage{booktabs}       
\usepackage{amsfonts}       
\usepackage{nicefrac}       
\usepackage{microtype}      
\usepackage{xcolor}         
\usepackage{algorithm2e}
\usepackage{graphicx}

\usepackage{hyperref}       
\usepackage{url}            

\title{High-Resolution Image Editing via\\ Multi-Stage Blended Diffusion}

\author{%
  Johannes Ackermann\thanks{Work done during an internship at Preferred Networks Inc.} \\
  The University of Tokyo\\
  \texttt{ackermann@ms.k.u-tokyo.ac.jp} \\
  \And
  Minjun Li \\
  Preferred Networks Inc. \\
  \texttt{minjunli@preferred.jp} \\
}

\begin{document}

\maketitle

\begin{abstract}
    Diffusion models have shown great results in image generation and in image editing. However, current approaches are limited to low resolutions due to the computational cost of training diffusion models for high-resolution generation. 
    We propose an approach that uses a pre-trained low-resolution diffusion model to edit images in the megapixel range.
    We first use Blended Diffusion to edit the image at a low resolution, and then upscale it in multiple stages, using a super-resolution model and Blended Diffusion.
    Using our approach, we achieve higher visual fidelity than by only applying off the shelf super-resolution methods to the output of the diffusion model. We also obtain better global consistency than directly using the diffusion model at a higher resolution.
\end{abstract}

\begin{figure}[h]
    \centering
    \includegraphics[width=\columnwidth]{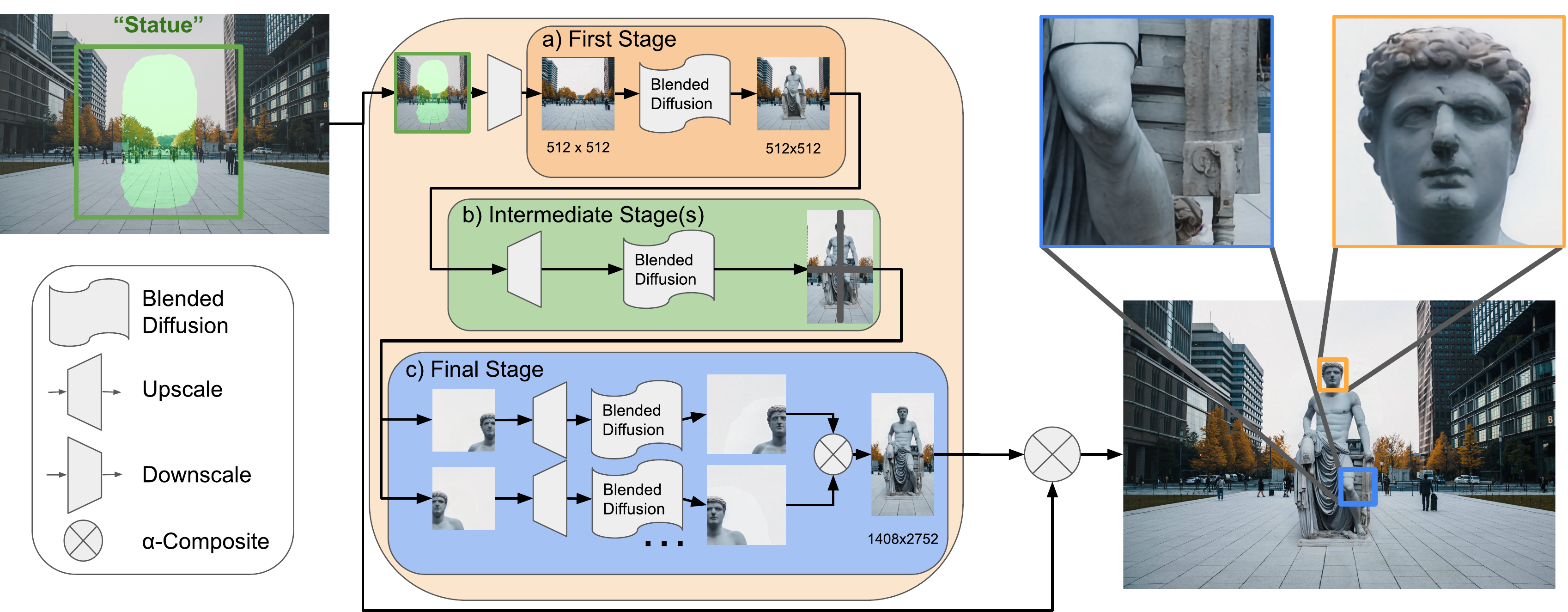}
    \caption{Our approach performs high-resolution text-guided image editing in multiple stages. In the first stage a), we apply Blended Diffusion \cite{avrahami2022blended}, given a masked region and a text prompt. In each following stage b), we first upscale the image using an off the shelf super-resolution model and then use Blended Diffusion, starting at an intermediate diffusion step, to improve the image quality and ensure consistency with the input prompt. c) When the output resolution of a stage is too large to fit into the GPU memory, we split the image into multiple segments, apply upscaling and Blended Diffusion to them separately and alpha-composite the results.}
    \label{fig:overview}
\end{figure}

\paragraph{Introduction}
Over the last year, along with great advances in text-guided image generation \cite{ramesh2022hierarchical,saharia2022photorealistic}, text-guided image editing approaches have been proposed with impressive results \cite{avrahami2022blended, nichol2021glide}.
Avrahami et al. \cite{avrahami2022blended} published a method called \textit{Blended Diffusion}, which allows us to reuse pre-trained text-guided diffusion models to edit images given a masked region and a text prompt.
However, as training large diffusion models is computationally expensive, publicly available models are limited to resolutions of at most 512x512 pixels.
Directly applying these models to higher resolutions leads to incoherent results, with repeated patterns and elements.
To overcome this issue, we propose a multi-stage approach that, by first editing the image at a low resolution and then upscaling the image in multiple stages, is able to edit large areas coherently at a high resolution.

\paragraph{Approach}
As illustrated in Fig. \ref{fig:overview}, we begin by cropping the image to a square area around the masked region and downscale this region to match the training resolution of the diffusion model.
In the first stage (Fig.\ 1a), we use Blended Latent Diffusion \cite{avrahami2022blended} to obtain a batch of five edited outputs.
We found it crucial to utilize Repaint \cite{lugmayr2022repaint} together with Blended Diffusion here, leading to a significantly higher consistency between the masked and unmasked region.
From the batch of results we select the best sample by similarity to the prompt using CLIP \cite{radford2021clip}, following related work \cite{avrahami2022blendedlatent, ramesh2021dalle}.
We then perform one or more intermediate stages (Fig.\ 1b) of upscaling using Real-ESRGAN \cite{wang2021realesrgan} followed by Blended Diffusion. We first take a forward step in the diffusion process to an intermediate timestep \cite{meng2021sdedit} and then denoise with Blended Diffusion.
As background input for Blended Diffusion, we found it helpful to not just add the same amount of noise as to the masked region, but also to first low-pass filter the masked region to match the spatial resolution of the edited region.
Without this preprocessing, the diffusion process tends to blur the region we are editing, as we show in the Appendix \ref{app:ablations}.
When the output resolution of a stage is too large (Fig.\ 1c), we divide the image into a grid of overlapping regions, processing each region separately. 
We then use alpha compositing \cite{porter1984compositing} to recombine the processed segments.
We also found performing decoder optimization \cite{avrahami2022blendedlatent} after each stage to significantly improve the blending between the edited region and the background.
\begin{figure}[b]
    \centering
    \includegraphics[width=\columnwidth]{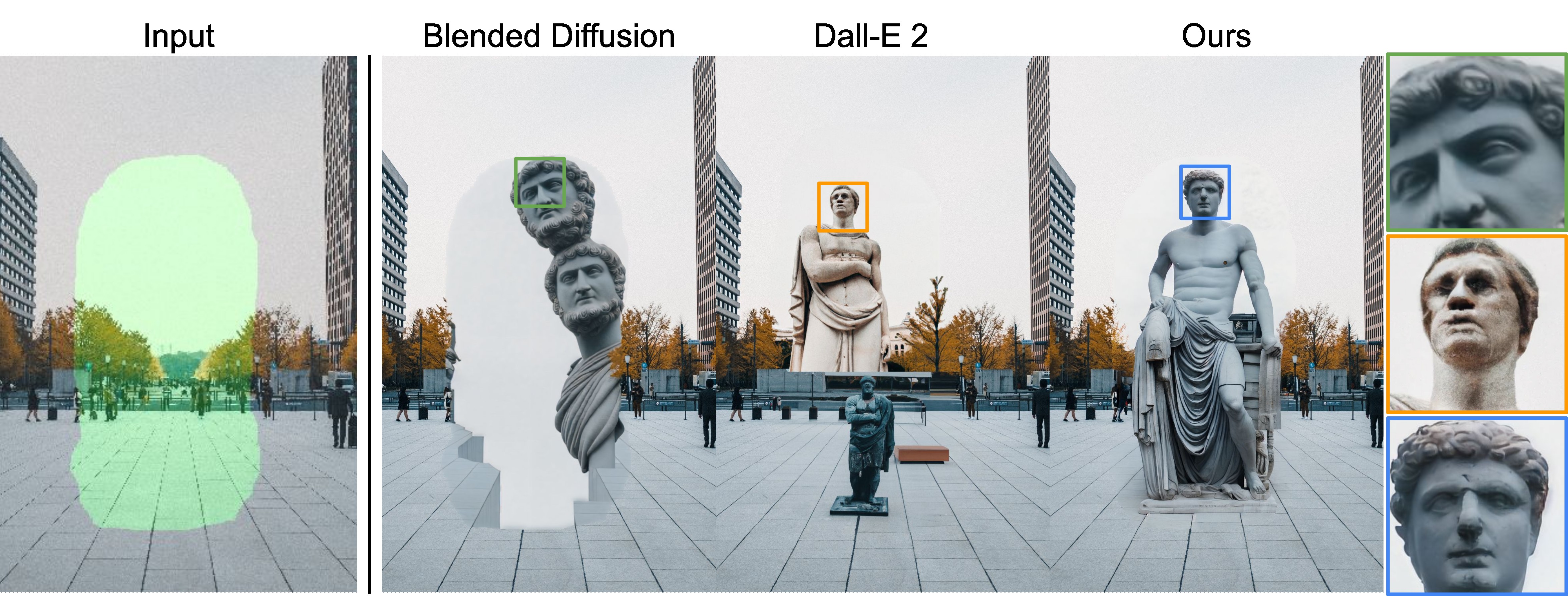}
    \caption{Comparison of our approach to two baselines for the prompt “Statue of Roman Emperor, Canon 5D Mark 3, 35mm, flickr”. From left to right: Blended Diffusion applied to 960x960 pixels followed by bilinear upscaling, Dall-E 2 editing in multiple segments, our proposed approach. The size of the mask is 1166x2297 pixels. Applying Blended Diffusion directly to the higher resolution leads to incoherent generation with repeated elements. Similarly, Dall-E 2 generates two statues, with one floating above the other. Our approach is able to generate a detailed, coherent image.}
    \label{fig:comparison}
\end{figure}

\paragraph{Evaluation}
We implement our approach based on the latent diffusion implementation provided by Rombach et al. \cite{rombach2022high} and use Stable Diffusion v1.4 \cite{rombach2022high}.
In Fig. \ref{fig:comparison}, we compare our approach to two baselines:
First, we directly apply Blended Diffusion \cite{avrahami2022blended} to the highest resolution we can fit in 32GB of VRAM, 960x960 pixels, and bilinearly upscale to the original resolution. 
Due to the resolution being higher than in training, the image is not globally consistent with multiple heads being generated.
As a second baseline, we use the editing function of the Dall-E 2 \cite{ramesh2022hierarchical} Web UI to paint in the masked region. 
As the Dall-E 2 model is limited to 1024x1024 pixels, we need to edit the image in six separate segments, leading to poor consistency between parts of the image. Two statues have been generated, one of which is floating above the other.
Finally, using our approach, we are able to produce a high-resolution result that is globally consistent.

We provide a discussion of the background and related work in the Appendix, as well as additional sample images, implementation details, ablations, and attribution of the used images.
Our source code is available at \url{https://github.com/pfnet-research/multi-stage-blended-diffusion}.

\paragraph{Ethical Impact}
Image editing, as well as image generation, can have significant ethical implications. While they can be used for many positive uses, for example for concept art generation, general image retouching or entertainment, they can also be used by malicious actors to fabricate images to pass off as real. However, at the current state it is possible with closer inspection to tell which images are edited. Another issue is that the pretrained model we use in our approach, Stable Diffusion v1.4 \cite{rombach2022high}, was trained on the LAION dataset, which is known to have significant biases with regard to ethnicity and gender. See Birhane et al. \cite{birhane2021multimodal} for an investigation of this issue.


\appendix

\section{Appendix}

\subsection{Background}
Diffusion models \cite{ho2020denoising} are a class of generative models, which define a forward process $q(x_t|x_{t-1})$ that adds noise to a given sample, such that the marginal distribution at timestep $t=T$ is a symmetric Gaussian distribution $p(x_T)=\mathcal{N}(x_T;\textbf{0};\textbf{I})$ and the marginal distribution at timestep $t=0$ is the data distribution that we aim to learn  $p(x_0) = p_D(x_0)$.
To generate samples from the data distribution, Denoising Diffusion Probabilistic Model (DDPM) \cite{ho2020denoising} learn the reverse process $p_\theta(x_{t-1}|x_t)$, which allows a sample to be generated by sampling $x_T \sim \mathcal{N}(\textbf{0};\textbf{I})$ and iteratively denoising it with $p_\theta(x_{t-1}|x_t)$. 

As the number of timesteps is usually large ($T\approx 1000$), each step requiring one forward pass of the network parametrizing $p_\theta$, both training and inference are computationally expensive.
Latent diffusion models (LDMs) \cite{rombach2022high} speed up training and sampling by splitting their approach into two stages: First, they train a variational autoencoder (VAE) with decoder $D_\theta$ and encoder $E_\theta$ to obtain a lower dimensional representation of the input images. In the second stage, they define the forward process in the latent space and train a DDPM model for the reverse process.
 
Blended Diffusion \cite{avrahami2022blended} provides a way to edit images with text-guided diffusion models. Given a mask $m$, an input image $x_{in}$ and text conditioning $c$, they alter the sampling process by combining the intermediate result $x_t$ with a noised version $\hat{x}_t$ of the input after each denosing step: $x_t \gets m \odot x_t  + (1-m) \odot \hat{x}_t$. $\hat{x}_t$ here is a sample of the forward diffusion process at time step $t$, starting from $x_{in}$; $\hat{x}_t \sim q(x_t|x_0=x_{in})$.
In a subsequent work, Avrahami et al. \cite{avrahami2022blendedlatent} apply Blended Diffusion to LDMs. They show that directly applying Blended Diffusion leads to poor blending with the original image due to the lossy encoding of the VAE. 
To address this issue, they propose to optimize the weights of the VAE decoder for each edited image to minimize the following loss
\begin{equation}
    L_{DO} = || m \odot (x_{out} - D_{\theta}(z_{out})) ||_{2}^2 + \lambda || (1 - m) \odot (x_{in} - D_{\theta}(z_{out})) ||_{2}^2 \,,
\end{equation}
where $z_{out}$ is the result of the editing in latent space, $x_{out} = D_\theta(z_{out})$ is the result in pixel space when using the original decoder weights, $x_{in}$ is the input image, and $\lambda$ weights the two loss terms.
In other words, this loss optimizes the unmasked area to be the similar to the unedited image and the masked area to be similar to the edited image.

SDEdit \cite{meng2021sdedit} is a method that allows unconditional or text-conditional diffusion models to be utilized as image-to-image models, projecting an input image $x_{in}$ onto the data distribution $p_D$ that a diffusion model $p_\theta$ was trained on.
To do so, they sample from the forward diffusion process at time $T' < T$, using $x_{in}$ as sample at time $t=0$, obtaining $x_{T'} \sim q(x_{T'} | x_0 = x_{in})$. They then use the trained diffusion model $p_\theta(x_{t-1}|x_t)$ to sample from the reverse process starting with $x_{T'}$, obtaining the result at $t=0$.

Repaint \cite{lugmayr2022repaint} is a method that utilizes unconditional diffusion models for inpainting. 
Starting from Gaussian noise at $t=T$, similar to Blended Diffusion \cite{avrahami2022blended}, after every denoising step they replace the unmasked area with a noised version of the input image. 
However, instead of sampling from the reverse process by going from step $t=T$ to $t=0$ linearly, they repeat each denoising step $R$ times, by taking forward and backward steps in the diffusion process iteratively. 
While this increases the runtime, they show that it also significantly increases the quality of the inpainting result.

\begin{figure*}
    \centering
    \includegraphics[width=\textwidth]{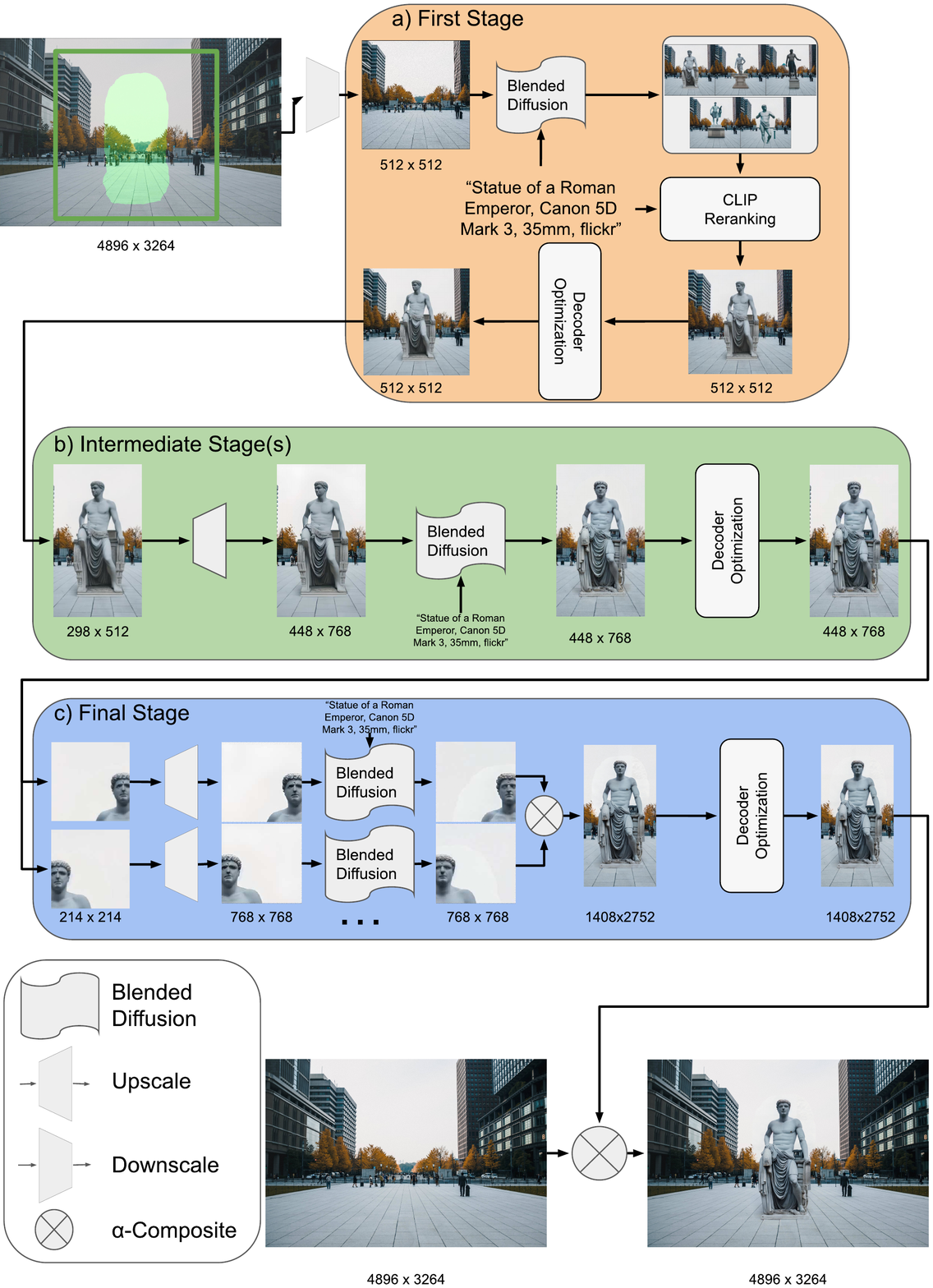}
    \caption{Our approach, with all intermediate results being shown. Best viewed zoomed in.}
    \label{fig:approachdetailed}
\end{figure*}

\SetKwComment{Comment}{$//$\ }{}
\SetKwInOut{Input}{Input}
\newcommand\mycommfont[1]{\footnotesize\ttfamily\textcolor{blue}{#1}}
\SetCommentSty{mycommfont}
\RestyleAlgo{ruled}
\DontPrintSemicolon
\LinesNumbered

\begin{algorithm}
\caption{Multi-Stage Blended Diffusion}
\label{alg:msbd}
\Input{input image $x_{in}$, mask $m$, text prompt $c$, DDPM $p_\theta$, forward process $q$, super-resolution model $\mathrm{SR}$, low-pass filter $\mathrm{LP}$, stage count $n_s$, stage-wise start timestep $T'_s$, repaint step $R$, first stage batch size $B$}

$\tilde{x} \gets \mathtt{Downsample}(\mathtt{Crop}(x_{in},m))$\;
$\mathcal{X}^1 \gets \emptyset$\;
\For{$i \in [1,\dots\,B]$} {
    $x_{T} \sim q(x_{T}|x_0=\tilde{x})$\;
    \For(\tcp*[f]{First stage, Blended Diffusion with Repaint}){$t \in [T,\dots,0]$ }{ 
        $x_{t-1} \sim p_\theta(x_{t-1}|x_t,c)$\;
        $\hat{x} \sim q(x_{t-1}|x_0=\tilde{x})$\;
        $x_{t-1} \gets m \odot x_{t-1} + (1-m)\odot \hat{x}$\; 
        \For{$r \in [1,\dots,R]$}{
            $x_t \sim q(x_t|x_{t-1}=x_{t-1})$\;
             $x_{t-1} \sim p_\theta(x_{t-1}|x_t,c)$\;
           $x_{t-1} \gets m \odot x_{t-1} + (1-m)\odot \hat{x}$\; 
        }
    }
    $\mathcal{X}^1 \gets \mathcal{X}^1 \cup {x_0}$\;
}
$x^1 \gets \mathtt{ClipReranking}(\mathcal{X}^1,c)$\;
$x^1 \gets  \mathtt{DecoderOptimization}(x^1,x_{in},p_\theta)$ \;
\For(\tcp*[f]{Intermediate stages, Blended Diffusion without Repaint}){$s \in [2,\dots,n_s-1]$}{
    $\tilde{x} \gets m\odot \mathrm{SR}(x^{s-1})  + (1-m) \odot \mathrm{LP}(x_{in})$\;
    $x_{T'_s} \sim q(x_{T'_s}|x_0=\tilde{x})$\;
    \For{$t \in [T'_s,\dots,0]$}{ 
        $x_{t-1} \sim p_\theta(x_{t-1}|x_t,c)$\;
        $\hat{x} \sim q(x_{t-1}|x_0=\tilde{x})$\;
        $x_{t-1} \gets m \odot x_{t-1} + (1-m)\odot \hat{x}$\; 
    }
    $x^s \gets \mathtt{DecoderOptimization}(x_0,x_{in},p_\theta)$
}
\Comment{Final stage processed in segments, omitted for brevity, see Fig. \ref{fig:approachdetailed}}

$x_{out} \gets m \odot x^{n_s} + (1-m)\odot x_{in}$

\end{algorithm}

\subsection{Implementation Details}
See Figure \ref{fig:approachdetailed} for a detailed diagram of our approach, with outputs of all intermediate processing steps being shown, and Algorithm \ref{alg:msbd} for an algorithm showing our approach.

We build our implementation based on the latent diffusion model (LDM) implementation by Rombach et al. \cite{rombach2022high} and use Stable Diffusion v1.4 \cite{rombach2022high} for our experiments.
However, our approach is also applicable to pixel-space diffusion models.
As outlined in the main text, given an input image $x_{in}$, a mask $m$, and a text conditioning $c$, we begin by cropping the image to a square region around the mask.
Because the diffusion model only receives the selected area as input, it is important that this area includes the context necessary for consistent inpainting. 
We, therefore, set a margin around the masked region for each example which ensures the necessary context is included. 
This area is then downsampled to the resolution the LDM was originally trained on.
In the first stage, we use a combination of Blended Diffusion \cite{avrahami2022blended} and Repaint \cite{lugmayr2022repaint}, as shown in lines 4-14 of Algorithm \ref{alg:msbd}.

We generate a batch of $B$ images and select the best one by similarity to the prompt using CLIP ViT/L14 \cite{radford2021clip}, following related work \cite{avrahami2022blendedlatent, ramesh2021dalle}.

We then perform multiple stages $s\in [2,..,n_s]$ of upscaling, consisting of applying a pre-trained super-resolution model $\mathrm{SR}$ and then Blended Diffusion with SDEdit \cite{meng2021sdedit}.
As $\mathrm{SR}$, we use "realesrgan-x4plus" \cite{wang2021realesrgan} followed by bilinear downscaling.
As the RealESRGAN model was trained to upscale natural images with natural distortions, artifacts caused by the diffusion model in the previous stage usually are present in the output of $\mathrm{SR}$.
Therefore, we then use Blended Diffusion to improve the visual fidelity and ensure consistency of the generation with the prompt.
As input $\tilde{x}$ for Blended Diffusion, we use the output of $\mathrm{SR}$, but replace the unmasked region with a low-pass filtered version of the input image: $\tilde{x} = m \odot \mathrm{SR}(x^{s-1}) + (1-m) \odot \mathrm{LP}(x_{in})$, where $\mathrm{LP}$ is a low-pass filter and $x^{s-1}$ is the output of the previous stage.
We discuss the reason for the filtering in more detail in Section \ref{sec:ablations:upscaling}.

We implement the low-pass filter $\mathrm{LP}$ by bilinearly downsampling the input image $x_{in}$ to the input resolution of the current stage and then upsampling it to the target resolution.
After each stage, we perform decoder optimization to improve the blending with the original image.

While we could, barring memory constraints, apply the diffusion model to arbitrarily large resolutions, in practice even a generous graphics memory of 32GB limits us to at most 960x960 pixel when using Stable Diffusion.
Therefore, for higher resolutions we split the input into multiple overlapping segments and apply the upscaling and Blended Diffusion to them separately.
We then combine the upscaled segments by alpha-compositing \cite{porter1984compositing} the overlapping regions.
Finally, as we also can not perform decoder optimization at high resolutions directly, we again split the image into multiple segments and optimize the loss $L_{DO}$ across all segments.
Using the optimized weights, we then decode each input segment and again use alpha compositing to combine them.

\paragraph{Hyperparameters}
In our experiments, we use one intermediate stage with the longer edge of the output being 768 pixels long, followed by the final full resolution which is processed in segments. The segments are each 768x768 pixels large overlapping with each neighbor by 128 pixels.
To speed up inference, we use DDIM \cite{song2021denoising} as sampler with $T=50$ steps.
In the first stage we generate a batch of size $B=5$ as input for the clip reranking.
In the intermediate stages we start the denoising from $T'=0.4T$ and in the final, segmented stage we start from $T'=0.25T$.
We run decoder optimization for 100 steps, using Adam \cite{kingma2014adam} with learning rate $1\times10^{-4}$, as suggested in \cite{avrahami2022blendedlatent}.
We use $R=5$ repaint steps in the first stage and don't use Repaint in the following stages.

\begin{figure}
    \centering
    \includegraphics[width=0.7\columnwidth]{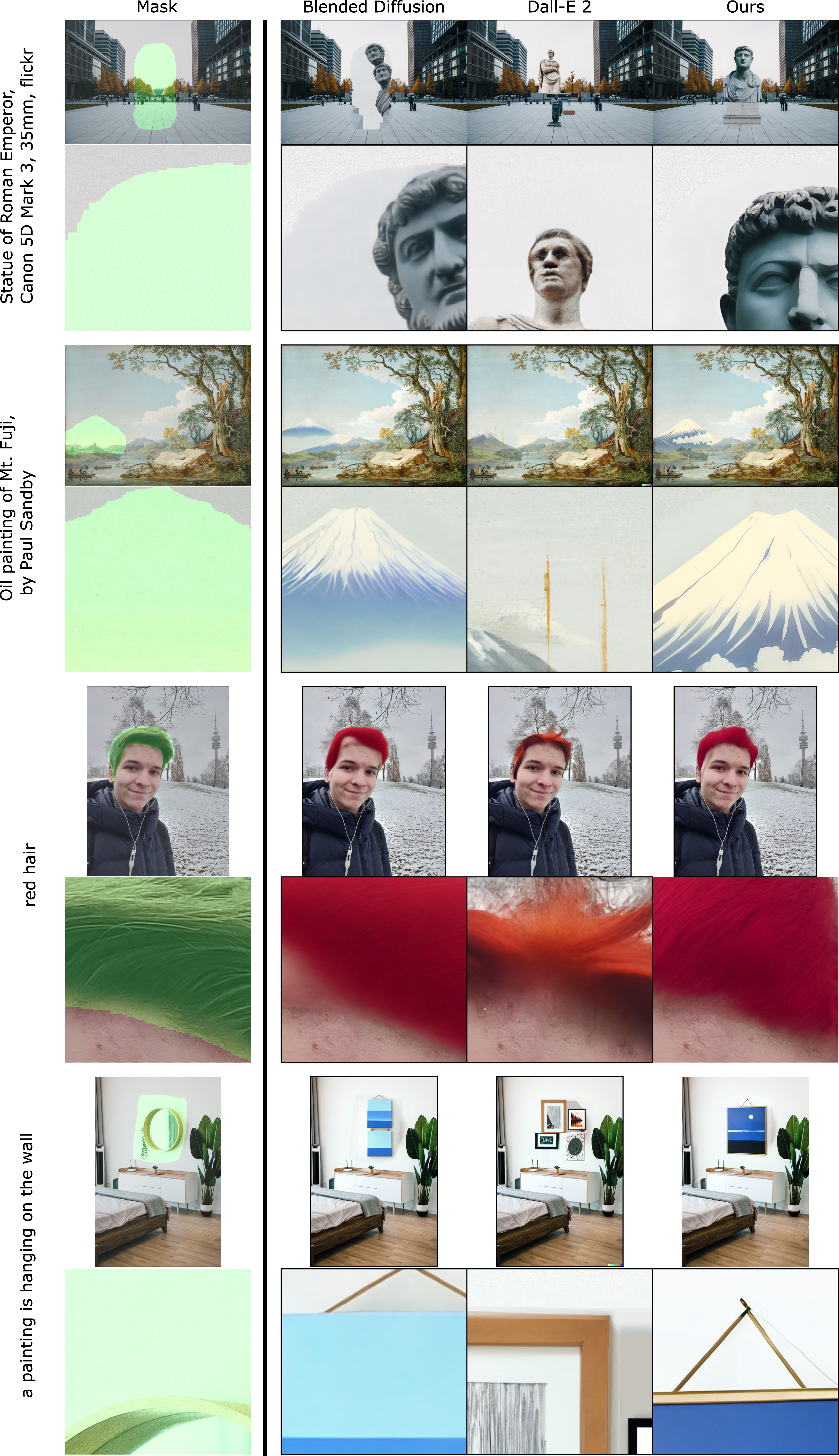}
    \caption{Comparison of our approach to two baselines. Left: we directly apply Blended Diffusion to the highest resolution we can fit into the VRAM (960x960 pixels) and then bilinearly upscale the output. Middle: We use the Dall-E 2 web UI to edit the image at its full resolution. Due to the edited region being larger than the 1024x1024 generation window, we have to apply Dall-E to multiple independent segments. Right: Our proposed approach. We find that directly applying Blended Diffusion leads to repeated elements (two heads, two mountains, two pictures) and fails to produce fine details (hair). Dall-E 2 produces visually high-fidelity images, but fails to produce globally coherent images (floating statues, four paintings). Our method produces globally consistent images while providing a similar visual fidelity. Note that the full images are downscaled. The zoomed-in regions measure 512x512 pixels and are shown at full resolution.}
    \label{fig:baselines}
\end{figure}

\subsection{Baseline Comparisons}
To validate our approach, we compare it to two baselines:
1) Using Blended Diffusion with Stable Diffusion v1.4 naively by running the first stage at the highest resolution we can fit in VRAM, 960x960 pixels in our case, and then upscaling the edited images bilinearly.
2) Using Dall-E 2 \cite{ramesh2022hierarchical} via their Web-UI, we can inpaint the masked region at high resolutions. However, as Dall-E 2 is limited to 1024x1024 pixels, we have to inpaint in multiple segments separately.
The results are shown in Figure \ref{fig:baselines}.
We find that directly applying Blended Diffusion to a resolution significantly higher than the model was trained on leads to repeated elements, such as the two mountains or two paintings in our examples.
Similarly, while producing a high-fidelity image, applying Dall-E 2 in segments also produced repeated elements (statues, paintings).
Our multi-stage approach allows us to obtain a globally consistent result.

\subsection{Ablations}
\label{app:ablations}

\begin{figure}
    \centering
    \includegraphics[width=\columnwidth]{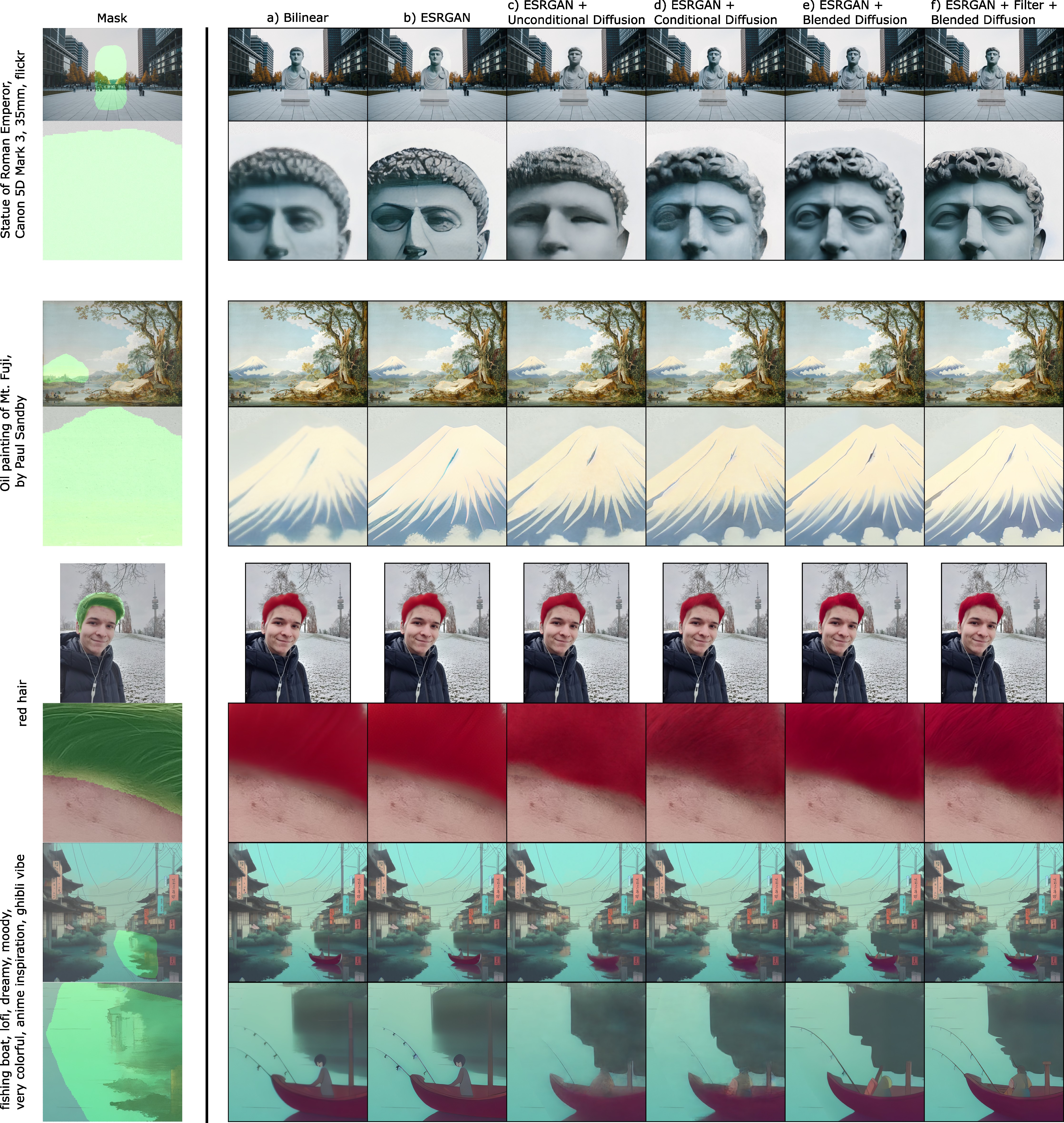}
    \caption{Ablation of different upscaling methods, applied after the Blended Diffusion in the first stage with fixed seeds. a) Bilinear upscaling, b) ESRGAN, c) ESRGAN + unconditional diffusion, d) ESRGAN + text-conditional diffusion, e) ESRGAN + text-conditional Blended Diffusion, f) ESRGAN + text-conditional Blended Diffusion with a low-pass filtered background. Note that the images are downscaled from the full resolution.}
    \label{fig:upscaling}
\end{figure}

\paragraph{Upscaling}
\label{sec:ablations:upscaling}
To upscale the image or image segments after the first stage, we upscale the low-resolution input with Real-ESRGAN and then use text-guided Blended Diffusion with a low-pass filtered background image. 
To validate our design decisions, we compare a) only bilinear upscaling, b) only ESRGAN upscaling, c) ESRGAN + unconditional diffusion, d) ESRGAN + text-conditional diffusion, e) ESRGAN + text-conditional Blended Diffusion with an unfiltered background image f) ESRGAN + text-conditional Blended Diffusion with a low-pass filtered background image. 
We show the results for the different ablations in Figure \ref{fig:upscaling}, showing that using our proposed method significantly improves blending with the background and the visual fidelity of the results.

We also note that if we do not low-pass filter the background, Blended Diffusion tends to produce more blurred outputs.
We hypothesize that this is due to the difference in "sharpness" between the high-resolution background image and the upscaled output of the previous stage.
This difference then seems to be amplified in the denoising process, leading to a more blurry edited region.
We stress that this is an empirical observation and more work is needed to properly understand this phenomenon.

\begin{figure}
    \centering
    \includegraphics[width=\columnwidth]{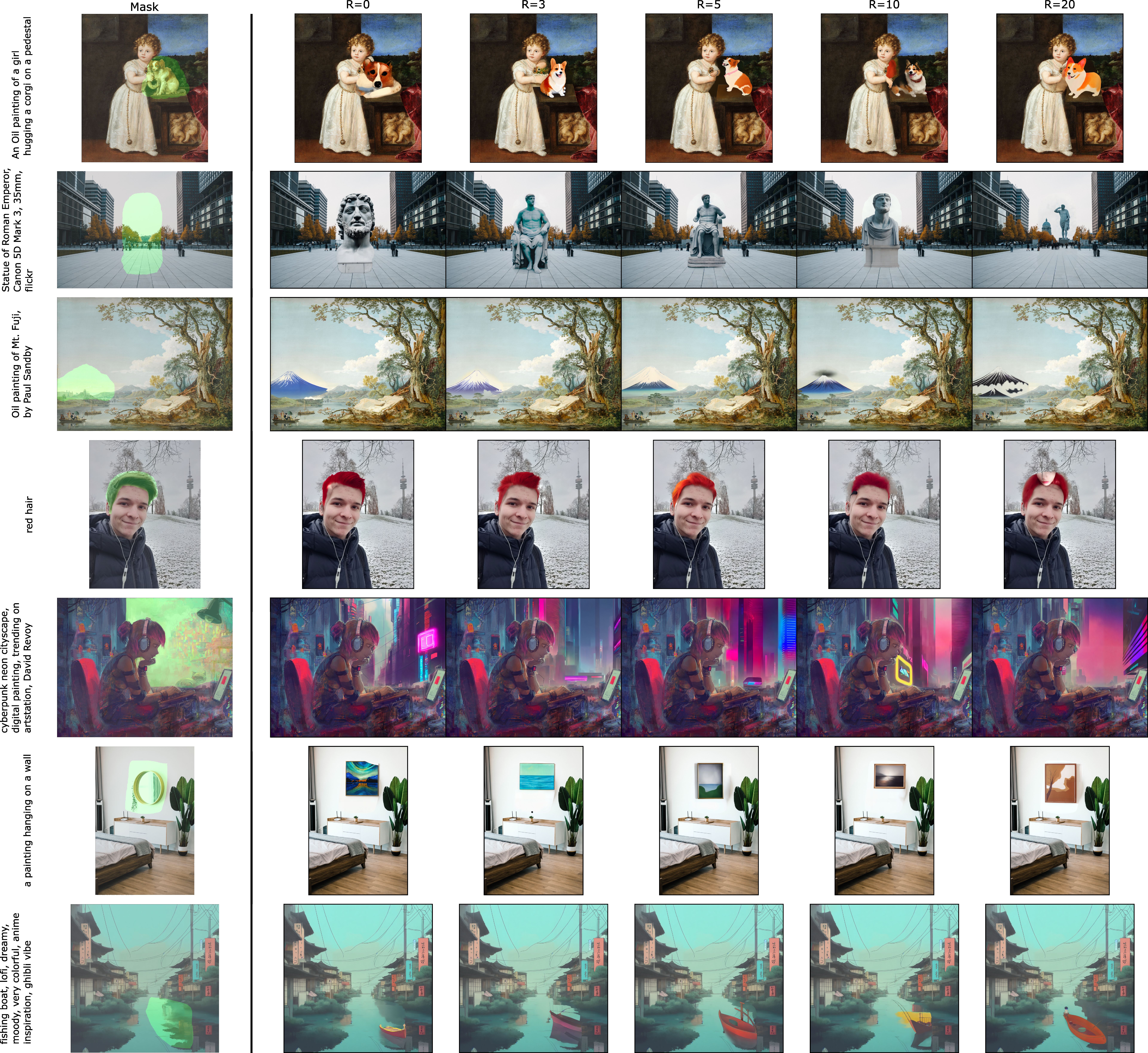}
    \caption{Comparison of different repaint steps $R$ without clip reranking. Without repaint, the edited region often does not blend well with the background, while too many repaint steps can also cause unrealistic images (corgi, hair). We generally find $R=5$ to work best. Note that the images are downscaled from the full resolution.}
    \label{fig:repaint}
\end{figure}

\paragraph{Repaint}
As we use Repaint \cite{lugmayr2022repaint} in the first stage of our approach, we provide an ablation of the effect of different repaint steps $R$ on the final result. We do not use clip-reranking in this experiment in order to isolate the effect of Repaint. The results are shown in Figure \ref{fig:repaint}. 
While not using repaint often leads to poor blending of the edited area, such as in the corgi, painting, and hair examples, using a too high number of repainting steps can also lead to poor results (hair, corgi).
Unlike the other samples, the river scene was generated by Stable Diffusion and editing it with Stable Diffusion seems to work well even without repaint.
In practice we find that setting the number of repaint steps to $R=5$ works well in most cases, however, a practitioner may want to adjust this parameter on a case-by-case basis to obtain the best possible results.

\begin{figure}
    \centering
    \includegraphics[width=\columnwidth]{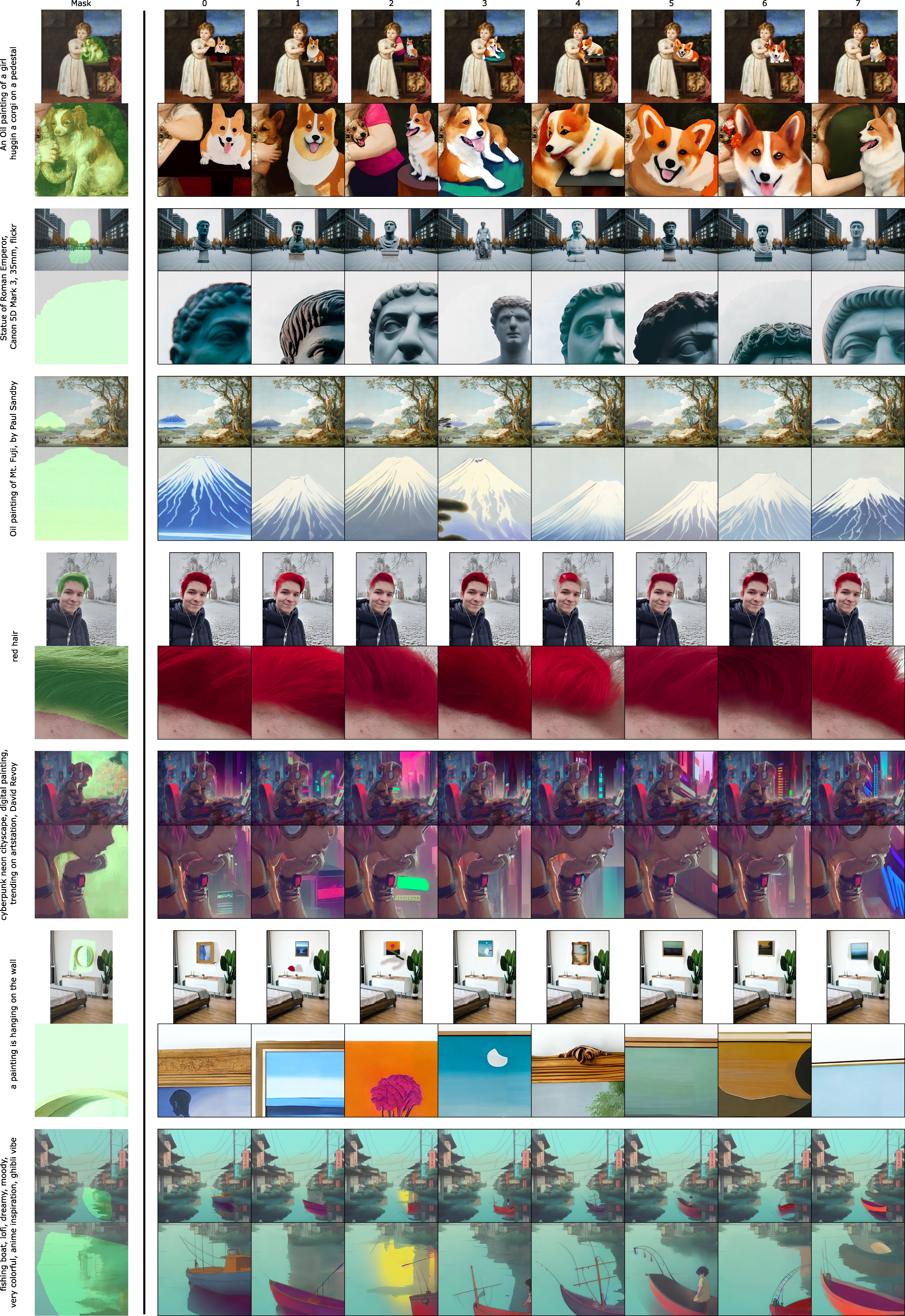}
    \caption{Outputs of our model for different random seeds. Note that the images are downscaled from the full resolution.}
    \label{fig:variations}
\end{figure}

\paragraph{Random Seeds}
To show the variation of our approach to random seeds we show results for different inputs for 8 different seeds each in Figure \ref{fig:variations}.
The quality of the results does vary between seeds, but for a practitioner it should be easy to run the model multiple times and pick the most appropriate output.

\subsection{Related Work}
Multi-stage approaches have been previously proposed for image generation \cite{cai2019multi,razavi2019vqvae2}.
Cascaded Diffusion \cite{ho2022cascaded} trains one model to generate images at a low resolution and then two super-resolution networks which are conditioned on the low-resolution input through channel-wise concatenation. A similar architecture is also utilized in Imagen \cite{saharia2022photorealistic} and DALL-E 2 \cite{ramesh2022hierarchical}.
Unlike these methods we use the same model in all stages and do not explicitly condition on the low-resolution image, allowing us to use any pre-trained text-conditional model. 
Additionally, these methods are in practice still limited to 1024x1024 pixels by computational constraints.
Wu et al. \cite{wu2022nuwainft} present an approach that is able to generate arbitrarily high-resolution images and perform  high-resolution outpainting by learning a representation of the nearby context, allowing global consistency.
However, their work is focused on out-painting or image generation and they do not investigate text-guided editing.
Glide \cite{nichol2021glide} presents an approach to generate and edit images using a text-conditional diffusion model. Similar to \cite{avrahami2022blended}, their approach is limited to the resolution that their diffusion model was trained on, and can not be easily extended to higher resolutions.

\section{Limitations}
Our current approach still has limitations, in three main ways:
Firstly, we reuse pretrained diffusion models and are therefore limited to editing images that they perform well at generating. For example, we found stable diffusion to perform poorly on night-scenes and in general it seems to not perform as well on photographs as it does on artwork.
Secondly, the blending between the unedited region and edited region is often still visible upon close inspection. Decoder optimization works well to improve the blending, but once we reach resolutions around 1000x2000 or above the optimization starts to take a significant amount of time (10 minutes on a V100 for 3000x2000 pixels).
Using a pixel-space diffusion model instead would remove the need for decoder optimization.
Thirdly, our approach uses Real-ESRGAN to upscale the images. While this works well in most cases, in some cases fine details are removed, leading to poor blending with the original image.

\section{Images used}
All images used in our paper are used in accordance with their licenses and attributed below, in the order of Figure \ref{fig:variations}:
\begin{enumerate} 
    \item "Portrait of Clarissa Strozzi", by Titian Vecelli, via \url{https://commons.wikimedia.org/wiki/File:Clarissa\_Strozzi,\_por\_Tiziano.jpg} (1803x2117)
    \item "people walking on sidewalk near high rise buildings during daytime", by Nat Weearwong, via \url{https://unsplash.com/photos/0cZgvYHirBg} (4896x3264)
    \item "The River Severn at Shrewsbury, Shropshire", by Paul Sandby, via \url{https://https://unsplash.com/photos/HEEvYhNzpEo} (3999x3041)
    \item Selfie by author (3456x4608)
    \item "white wooden dresser with mirror photo", by Minh Pham, via \url{https://unsplash.com/photos/7pCFUybP\_P8} (3902x5853)
    \item "Lofi Cyberpunk" by David Revoy \url{https://www.davidrevoy.com/article867/lofi-cyberpunk} (2431x1930), CC BY 4.0 \url{https://creativecommons.org/licenses/by/4.0/}
    \item Anime-style image of river generated with stable diffusion by the authors (2048x2048)
\end{enumerate}

\end{document}